\documentclass{article}

\usepackage[final]{neurips_2024}

\usepackage[utf8]{inputenc} 
\usepackage[T1]{fontenc}    
\usepackage{url}            
\usepackage{booktabs}       
\usepackage{amsfonts}       
\usepackage{nicefrac}       
\usepackage{microtype}      
\usepackage{xcolor}         
\usepackage{todonotes}
\usepackage{tcolorbox}
\usepackage{mathmacros}
\usepackage{mathtools}
\usepackage{enumitem}
\usepackage{amsthm}
\usepackage{algorithm}
\usepackage{algorithmic}
\usepackage{titletoc}
\usepackage{wrapfig}
\usepackage{pifont}
\usepackage{authoraftertitle}
\usepackage{wrapfig}
\usepackage{xurl}
\usepackage{zi4}

\theoremstyle{plain}

\newtheorem*{theorem*}{Theorem}

\newtheorem*{proposition*}{Proposition}

\newtheorem*{corollary*}{Corollary}
\newtheorem{definition}{Definition}

\theoremstyle{remark}

\numberwithin{equation}{section}

\title{Algorithmic Phase Transitions in Language Models: \\ A Mechanistic Case Study of Arithmetic}

\author{%
  Alan Sun$^1$, Ethan Sun$^2$, Warren Shepard$^2$\\
  $^1$Carnegie Mellon University, $^2$Dartmouth College \\
  \texttt{awsun@cmu.edu, \{ethan.k.sun, warren.a.shepard\}.26@dartmouth.edu }
}

\usepackage[hidelinks]{hyperref}
\hypersetup{
    colorlinks,
    linkcolor={red!50!black},
    citecolor={blue!50!black},
    urlcolor={blue!80!black}
}

\begin{document}

\maketitle

\begin{abstract}
Zero-shot capabilities of large language models make them powerful 
tools for solving a range of tasks
without explicit training. It remains unclear, however, how these 
models achieve such performance, or why 
they can zero-shot some tasks but not others. In this paper, we shed some 
light on this phenomenon by 
defining and investigating \textit{algorithmic stability} in language models
---changes in problem-solving
strategy employed by the model as a result of changes in task specification. We 
focus on a task where algorithmic stability is needed for generalization: 
two-operand arithmetic. Surprisingly, we find that \texttt{Gemma-2-2b} 
employs substantially different computational models 
on closely related subtasks, i.e. four-digit versus eight-digit addition. 
Our findings suggest that algorithmic instability may be a contributing factor 
to language models' poor zero-shot performance across certain logical reasoning tasks, 
as they struggle to 
abstract different problem-solving strategies and smoothly transition 
between them\footnote{Codebase for our experiments can be found at
\url{https://github.com/alansun17904/mech-arith}}.
\end{abstract}

\section{Introduction} 
Language models have demonstrated remarkable capabilities across many benchmarks. 
In particular, their zero-shot performance---solving a task or generating
outputs for which they have not been explicitly trained on---makes them powerful tools.
However, it remains unclear how this behavior emerges from training against the language 
modeling objective~\citep{wei_emergent_2022, schaeffer_are_2023}. Further, 
it has been shown that zero-shot performance varies highly between 
different tasks~\citep{weichain2022}. In this paper, we shed light
on this phenomenon by exploring the robustness of a models' problem solving
strategy with respect to changes in its given task. For example,
given two-operand addition problems, we would expect a \textit{strong learner}
to add four-digit numbers the same way it adds eight-digit numbers. We 
coin the term \textit{algorithmic stability} to describe this desirable
property (see Def.~\ref{def:alg-stable}). Surprisingly, we find that even
for a small language model like \texttt{Gemma-2-2b}~\citep{gemmateam2024gemma2improvingopen}, substantially different
computational models are implemented on closely related subtasks. Our findings
suggest that algorithmic instability may be a contributing factor to language models'
poor zero (few)-shot performance across certain logical reasoning tasks, as they
struggle to abstract different problem-solving strategies and smoothly transition
between them. 

We call these aforementioned changes in problem solving strategies under the same task
\textit{algorithmic phase transitions} (see Def.~\ref{def:alg-phase}). These phases are identified using mechanistic interpretability. Mechanistic
interpretability causally attributes~\citep{vig_investigating_2020, nanda_progress_2022, olsson2022context,chughtai_toy_2023} and localizes behavior to individual model components. The goal is to identify a minimal computational subcircuit that completely characterizes the model's
decision making process on a specific task~\citep{wang_interpretability_2022}. We identify subcircuits across a host of arithmetic subtasks, and then compare them both
quantitatively and qualitatively. 

Seminal works, such as~\citep{nanda_progress_2022, zhong_clock_2023}, have shown that Transformers~\citep{vaswani_attention_2017} implement different algorithms in tandem to solve the same task. Additionally, the interactions between these parallel algorithms are highly sensitive to training and architectural hyperparameters. Similarly,~\cite{merullo_circuit_2023} presented a case study of circuits within the same model being reused to perform two distinct tasks. Our work builds off of these results. We explore \texttt{Gemma-2-2b}'s algorithmic stability on two-operand arithmetic. Although arithmetic is a relatively elementary task, it represents a complex task for language models, especially when the operands have many digits~\citep{hendrycks_measuring_2021,hanna_how_2023, mao_wamp_2023, qiu_dissecting_2024}. 
The task encompasses sufficiently rich algorithms, since there are many possible implementations that span a host of problem-solving strategies, such as look-up table, divide-and-conquer, or dynamic programming. The methods described in our paper, however, can also be 
applied more broadly to other applications.

We first identify a collection
of algorithms implemented by the model to solve arithmetic. Then, we isolate sufficient
conditions in terms of task perturbations that cause the model to transition between
these algorithms. 
Moreover, we show that these phase transitions are induced by changes in subtask 
complexity. For example, as the number of digits in each operand increases, we present preliminary evidence that shows \texttt{Gemma-2-2b} undergoes periods of algorithmic stability and experiences sharp phase transitions. Our paper stands
in contrast to many existing works that seek to elucidate the mechanism for arithmetic
\citep{stolfo-etal-2023-causal, stolfo-etal-2023-mechanistic, qiu_dissecting_2024}, as we focus solely on the transitions between mechanisms rather than the mechanisms
themselves.

In sum, our contributions are threefold. Firstly, we introduce the notions of
\textit{\textbf{algorithmic stability}} and \textit{\textbf{algorithmic phase
transitions}}. Secondly, we
present evidence that for two-operand arithmetic, \textbf{\texttt{Gemma-2-2b} \textit{undergoes sharp algorithmic phase transitions.}} This may provide a sufficient explanation for their lack of generalization across arithmetic tasks. Lastly, we devise \textbf{\textit{novel methods and techniques to quantify these changes}} in contrast to many of the existing qualitative techniques of mechanistic interpretability.

\section{Defining Algorithmic Stability and Phase Transitions}
Herein, we first define the notions of \textit{algorithmic stability}
(see Def.~\ref{def:alg-stable}) and
\textit{algorithmic phase transitions} (see Def.~\ref{def:alg-phase}).
Next, we present methods to empirically identify phase transitions. Lastly,
we apply these methods to two-operand arithmetic and demonstrate how these
transitions can be used to explain model performance.

\begin{definition}[Algorithmic Stability]\label{def:alg-stable}
Let $T$ be a collection of tasks. Let $m$ be a model and denote 
$\alg(m, t)$ for $t \in T$ the minimal subcircuit of $m$ that explains $t$
~\citep{wang_interpretability_2022}. Then, we say that $m$ is $\epsilon$-algorithmically 
stable, for some $\epsilon > 0$, across $T$ with respect to some metric $\rho$ if
\begin{equation}\label{eq:alg-stable}
\sup_{t, t' \in T} \rho(\alg(m,t'), \alg(m, t)) \leq \epsilon.
\end{equation}
\end{definition}
Informally, if a model implements significantly different subcircuits across tasks
in $T$, we say that the model is algorithmically unstable over $T$. 
We also say that $m$ is \textit{algorthmically unstable} if Eq.~\ref{eq:alg-stable}
does not hold. So far, we have framed algorithmic stability as a desirable property.
But, depending on the composition of $T$, stability may be a hindrance to 
generalizability. For example, if $T$ contains $t_1$, the set of all addition problems,
and $t_2$, the set of all division problems. It would be necessary for performance
to solve both of these tasks with different \textit{minimal} circuits. 
In this paper, we focus on the case where algorithmic stability is desirable. 
Specifically, we define $T$ to be the set of all two-operand addition problems. 
Each $t \in T$ is a class of $m$-digit by $n$-digit addition problem, for
all $m, n \in \{1,2,\ldots, 8\}$. Therefore, $|T| = 64$. $\rho$ and $\alg(m,\cdot)$ is
defined through activation patching, which we describe in later subsections. 

If a model is algorithmically unstable on a set of tasks $T$, then we seek to identify \textit{algorithmic
phase transitions} between these tasks. That is, sufficient conditions in task specification that 
induce changes in subcircuitry. This notion is formally defined in Def.~\ref{def:alg-phase}.
\begin{definition}[Algorithmic Phase Transition]\label{def:alg-phase}
Let $m$ be a model and suppose that for a set of tasks $T$, metric $\rho$,
$m$ is $\epsilon$-algorithmically unstable, then $m$ undergoes an algorithmic
phase transitions between some $t, t' \in T$ if 
\begin{equation}
\rho(\alg(m, t'), \alg(m, t)) \geq \epsilon.
\end{equation}
\end{definition}
By Def.~\ref{def:alg-stable}, if $m$ is algorithmically unstable over $T$, 
then $m$ must undergo at least one algorithmic phsae transition in $T$.  

\begin{figure}
\centering
\makebox[0pt]{
\includegraphics[width=1.1\linewidth]{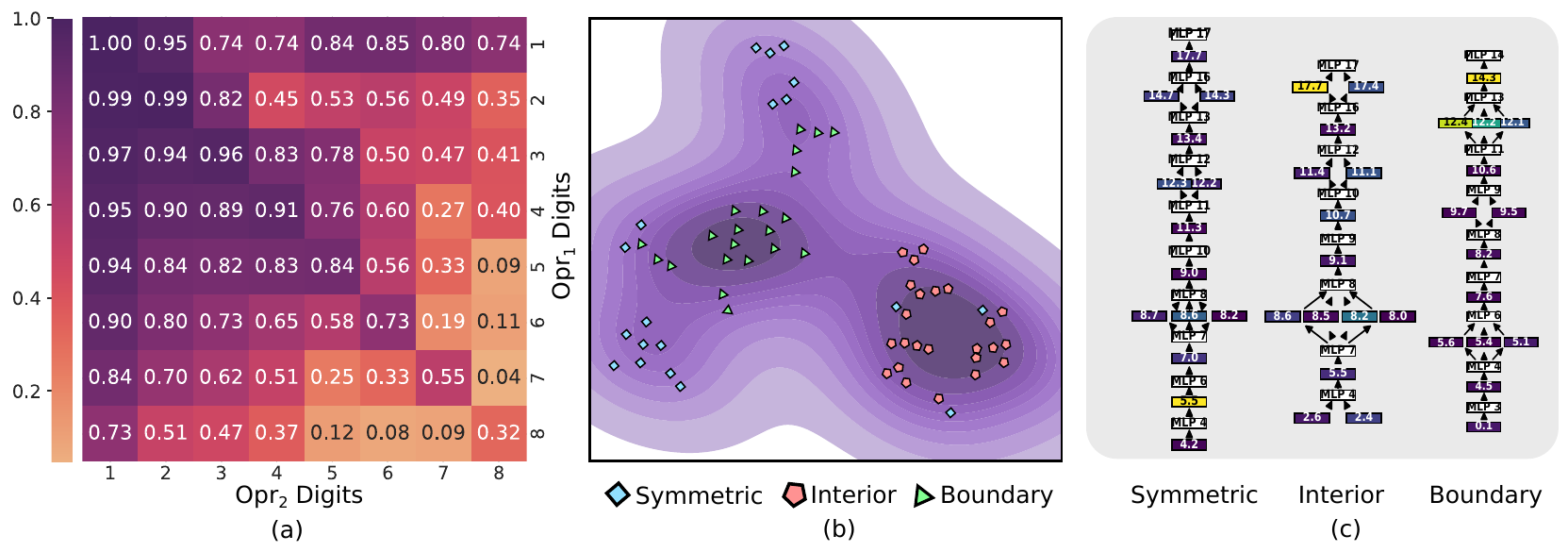}}
\caption{(a) Accuracy of \texttt{Gemma-2-2b} on two-operand addition as 
the number of digits in both operands vary. Accuracy is measured through 
exact string match. (b) t-SNE plot (with perplexity=3) where each point 
is the algorithm \texttt{Gemma-2-2b} implements for a $m,n$-digit addition 
problem. We capture algorithmic similarity measuring the importance of 
each attention head. (c) Sample circuits from $8,1$-, $6,3$-, and $4,4$-digit 
addition, from left to right. These tasks are representative of the phases 
we identify: symmetry, interior, and boundary.}
\label{fig:main-panel-cover}
\end{figure}

\section{Empirical Results}\label{sec:preliminaries}
All of our subsequent experiments are performed with \texttt{Gemma-2-2b}
\citep{gemmateam2024gemma2improvingopen}
on two-operand addition. We specifically chose arithmetic because we expect 
a model that performs well to learn general rules and algorithms. In subsequent sections, 
we reveal that this is not the case for \texttt{Gemma-2-2b}.

We consider problems of the form $\opra + \oprb$, where $\opra, \oprb \in \Z^+$ and $\opra,\oprb \leq 10^9 - 1$. Let us denote the class of $m,n$-digit addition problems as
the set $\{(a,b): 10^{m-1} \leq a \leq 10^m - 1, 10^n \leq b \leq 10^n -1\}$. In other words, $\opra, \oprb$
have $m,n$ digits, respectively. We aim to characterize the algorithm implemented by the model for any 
$m,n$-digit addition problem and then observe the differences as both $m,n$ change. In all of our
experiments, when we prompt a pretrained causal language model (in this case, \texttt{Gemma-2-2b}),
for a given problem $a + b=$, we first pass it $k=2$ few-shot examples from problems of the same 
class (same number of digits in each operand). An example prompt is shown below.
\begin{tcolorbox}
\vspace{-0.2cm}
\begin{align*}
\texttt{15 + 85 } &\texttt{= 100} \\
\texttt{65 + 12 } &\texttt{= 77} \\
\texttt{43 + 90 } &\texttt{=} 
\end{align*}
\end{tcolorbox}
We performed ablations over this hyperparameter, but observe marginal changes in 
performance as $k$ increases past $k=2$. 
Let us also denote the number of digits in $\text{Opr}_i$ as $|\text{Opr}_i|$. 

\subsection{Baseline Performance}
We benchmark \texttt{Gemma-2-2b} across all problem classes (all combinations of $m,n$-digit 
addition problems where $m,n \in \{1,2,\ldots,8\}$), without any intervention. 
For every problem class, we sample $n=1000$ problems uniformly at random. 
These results are shown in Fig.~\ref{fig:main-panel-cover}(a). 
We see that as the task complexity---the number of digits in both operands---increases, the performance of the model monotonically decreases. 
Moreover, the performance of the model is not symmetric. 
That is, the performance of $a + b$ is different from $b+a$ with a positive performance bias towards 
$|\opra| > |\oprb|$. Moreover, the magnitude of asymmetry increases as the number of digits 
in both operands increase. Notice also that the boundaries---where either $|\opra| = 1$ or $|\oprb| = 1$---of the matrix in Fig.~\ref{fig:main-panel-cover}
do not obey this property. Its performance does not degrade as quickly as
the number of digits in each operand increases, nor does it change as much as the number of digits change. 

\textbf{\textit{Based on these initial observations, we hypothesize three algorithmic
phase transitions defined by these aforementioend sharp changes in performance.}} 
More specifically, we partition all two-operand addition tasks into three sets: symmetric, boundary,
and interior. 
\begin{definition}[Task Partitions]
Consider the set of all arithemtic problems in the form of $\opra + \oprb = $. We say that 
a problem is symmetric if $|\opra| = |\oprb|$. A problem is on the boundary if 
either $|\opra| = 1$ or $| \oprb| =1$. And lastly, a problem on the interior is any 
task that does not fall into either of these categories. 
\end{definition}
Denote the set of symmetric, boundary, and interior problems by 
$S, B, I$, respectively. Clearly, if $T$ is the set of all two-operand addition problems,
then $T = S \cup B \cup I$. Across these sets, we see that as the model transitions between them, 
performance changes drastically. Using our definitions, we hypothesize that \texttt{Gemma-2-2b} is
algorithmically unstable across $\{S, B, I\}$ and undergoes algorithmic phase transitions
between all pairs of this set. We test this hypothesis by reverse engineering the subcircuits
used by the model for each subtask. 

\subsection{Reverse Engineering Arithmetic Circuits}
To find the causal subcircuits responsible for the algorithmic behavior 
of the model, we used \textit{activation patching}.  See \cite{heimersheim_how_2024} 
for a detailed treatment
of activation patching. We reverse engineer the subcircuits of each subtask independently. 
Let $T_{m,n,k}$ be the set of all $m,n$-digit addition problems with $k$ few-shot
examples. We proceed with denoising activation patching~\citep{meng_locating_2022,
wang_interpretability_2022, hanna_how_2023, merullo_circuit_2023}. 

Fix some $(x,y) \in T_{m,n,k}$ where $x$ is the input to \texttt{Gemma-2-2b} and $y$ is the
correct solution. We first perform a ``clean run:'' passing $x$ through to the model and storing all of the intermediate activations. Then, we perform a corrupted run: sample some other $x' \in T_{m,n,k}$ uniformly at random such that $x' \neq x$ which also has answer $y' \neq y$. In other words, we want to find an input such that its expected output should be different from the clean input. Now that we have the activation caches for both the clean and corrupted run, to determine if a computational component in the model has a sufficient causal impact on the performance of the model, we start by running the corrupted input through the model and then, at the component of interest, we patch in the output of that component from the clean run. Then, for every subsequent computational component, we recompute its output. Note that we only patch the outputs of the attention heads and modify the outputs of the MLPs. This is because the prevailing sentiment is that the attention heads are responsible for any algorithmic manipulations of the tokens that the model is doing, whereas the MLPs are mainly for knowledge and fact retrieval~\citep{geva_transformer_2021, geva_transformer_2022, haviv_understanding_2023}. Also, when we perform patching, we are patching the entire activation from the clean input into the model.

Let $y^\ast$ denote the model prediction with the replaced components. To quantitatively measure the impact of this computational component, we measure $\PP(y^\ast = y | x') - \PP(y^\ast = y | x)$. This is estimated by averaging the softmax of the target tokens in the logit layer across all of the token indices that we are predicting. 

We perform activation patching over $n=1000$ clean and counterfactual inputs for each task. The resulting circuits for all of the tasks can be found in Appendix~\ref{sec:circuit-additional}. We also present a detailed analysis of the different classes of circuits we find in the following subsections.

\subsection{Circuit Stability Under Across Task Classes}
Now that we have found the circuits, we want to see whether there is agreement in the set of components we have found within the proposed task classes we identify. As a result of activation patching, we have, for each attention head, the amount of performance we recover by patching in the clean output associated with it. Thus, using t-SNE, we compare these heatmaps by projecting it onto $\R^2$. The result of this is shown in Fig.~\ref{fig:main-panel-cover} (b). We see that algorithmic stability is generally conserved, but between the different sets we identified, the algorithms are drastically different. Some sample circuits within each of these classes are shown in Fig.~\ref{fig:main-panel-cover} (c).

\begin{figure}
    \centering
    \includegraphics[width=0.8\textwidth]{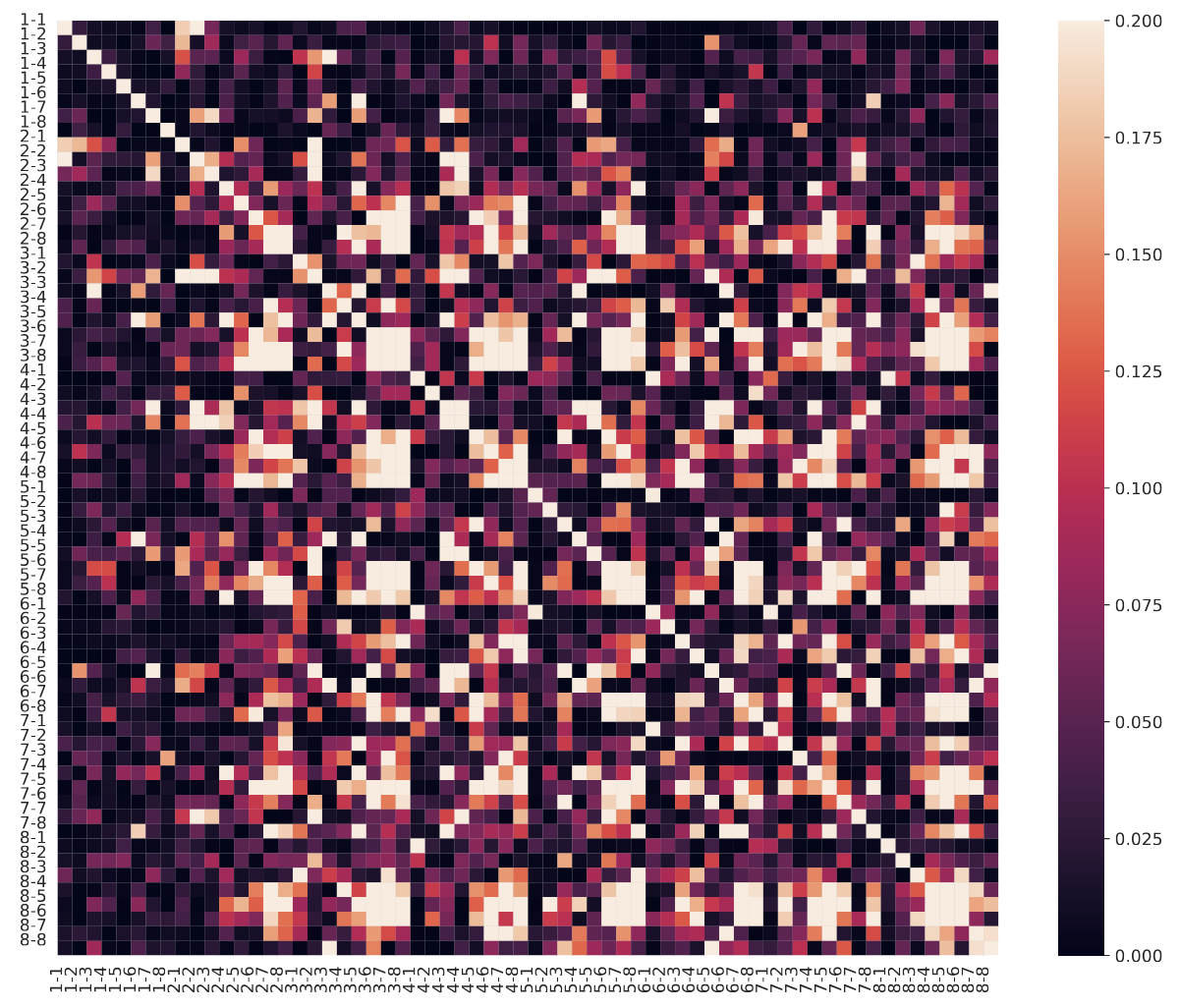}
    \caption{For a fixed model, \texttt{Gemma-2-2b}, this illustrates the pairwise-Pearson correlations between the attention head contributions found through activation patching. Each cell intuitively captures the pairwise subcircuit similarity between subtasks (see the $x,y$-axes labels).}
    \label{fig:pearson-corr}
\end{figure}

We also plot the Pearson correlation between the influence activations between every pair of tasks. This is shown in Fig.~\ref{fig:pearson-corr}. Though these plots demonstrate the existence of phase transitions across tasks for the model, it remains to characterize these phases algorithmically and reverse engineer the specific algorithms being applied in each of them.

\subsection{Characteristics of Phases}

\begin{figure}
    \centering
    \makebox[0pt]{
    \includegraphics[width=1.1\linewidth]{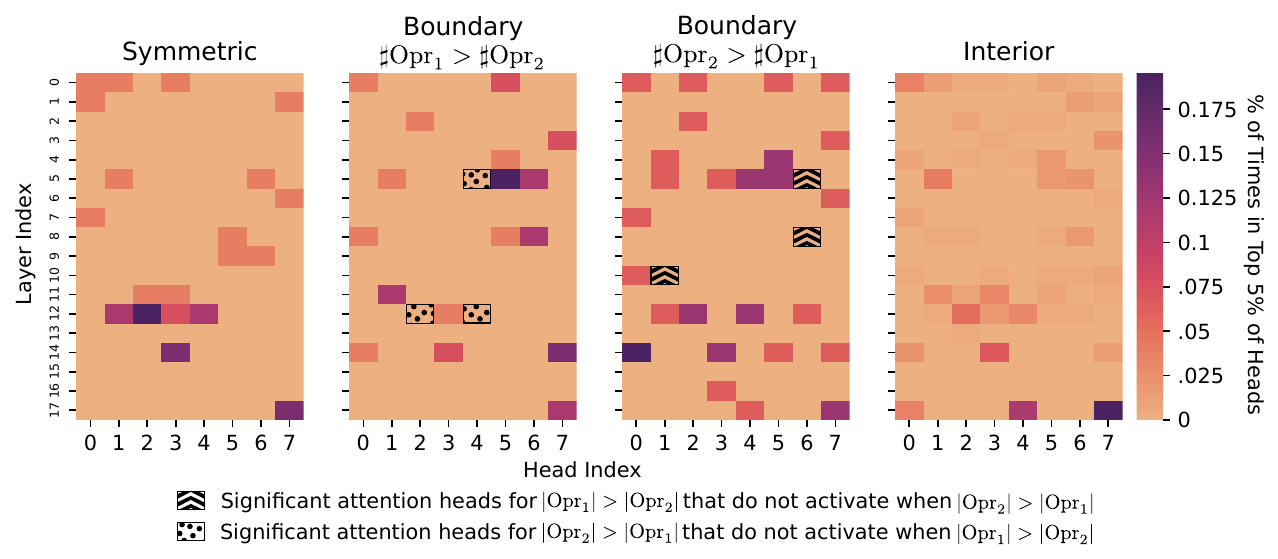}}
    \caption{For each subtask, each heatmap represents the probability that that particular attention head will be in the top 5\% of the most influential attention heads. We split the boundary subtask into two cases: where $|\opra| > |\oprb|$ and vice versa.}
    \label{fig:circuit-breakdown-heatmaps}
\end{figure}

Herein, we characterize the different algorithmic classes we claim to have identified based on their circuit diagrams. Please see Appendix~\ref{sec:circuit-additional} to see all of the circuits that we discovered. Note that to determine the underyling circuits, we looked at the top 10\% of all attention heads based on their influence score. 

We find that all of the subcircuits are activating in layer 0, specifically attention head 0 and also in the last layer at attention head 7. We posit that this interaction is responsible for task identification and also writing the final answer to the output stream. 

\textbf{Symmetric.} The $1,1$-digit addition circuit is different from all of the other circuits. This specific circuit is shown in the appendices. We hypothesize that this circuit is performing retrieval and memorization. Moreover, all of the algorithms for the phases that we identify below ``call'' this subcircuit as a recursive function. 

It can be observed that most of the circuits that fall into this classification are top-heavy. That is, most of their influential attention heads lie in layers 14-17. Moreover, there are also activations concentrated about the middle layers of the network, mostly layers 5 and 8. Specifically, we see that across all of the circuits, attention heads 12.2, 12.4, and 17.7 are activating strongly. There are also early activations in the first layer across heads 0.0, 0.2, and 0.3. Please refer to Fig.~\ref{fig:symmetric-circuits} to see all of the circuits associated with this subtask. This is shown in Fig.~\ref{fig:circuit-breakdown-heatmaps}.


\textbf{Boundary.} In contrary to the circuits that we found in the symmetric subtasks, we found that there are generally two peaks of activations: in the early layers and in the later layers. Specifically, when $|\opra| > |\oprb|$, we find that generally attention heads in layer 5 (5.5 and 5.6) are strongly activated, as well as attention heads in layer 14 (14.3 and 14.7). Similar to the circuits when $| \opra| = |\oprb|$, we see that attention heads in the early layers are strongly activated, such as 2.2, as well as the same attention heads in the first layer of the previous circuits. 

We also notice that within the boundary class, there seems to be two distinct behaviors. This is seen in Fig.~\ref{fig:main-panel-cover}. When $| \opra| > | \oprb|$ we see different circuits compared to $|\oprb| > | \opra|$. The subcircuits and their differences are further highlighted in Fig.~\ref{fig:circuit-breakdown-heatmaps}. We find that much more attention heads are involved in the computation of $|\oprb| > |\opra|$ case compared to the $| \opra| > | \oprb|$. Moreover, there are high-impact attention heads in the middle layers that are disjoint from each other. These attention heads are highlighted in Fig.~\ref{fig:circuit-breakdown-heatmaps}.

\textbf{Interior.} Lastly, circuits on the interior are generally diffuse throughout the model. This can be observed in the right-most panel of Fig.~\ref{fig:circuit-breakdown-heatmaps}. We find that, in general, many of the attention heads in the middle layer that are shared between the symmetric case and boundary cases can be found as a subset of the high-impact attention heads in this subtask. This suggests that algorithmically when \texttt{Gemma-2-2b} is solving problems belonging to the interior, it is using subcircuits from the boundary and symmetric cases in tandem.

\section{Discussion}
In this paper, we explored algorithmic phase transitions induced by changes in task complexity. Surprisingly, we found that even on simple tasks such as arithmetic, language models exhibit sharp algorithmic phase transitions across different subtasks. Our work presents a novel method to operationalize and quantify this transition by identifying and comparing the structures of the underlying circuits driving these behaviors. 

Our work relies on a host of seminal techniques in mechanistic interpretability, such as those by ~\cite{wang_interpretability_2022, hanna_how_2023, power_grokking_2022}. Specifically, we heavily rely on activation patching to identify these subcircuits. The limitations of activation patching are known and discussed extensively throughout the literature. However, they still remain the most effective methods to tractably identify subcircuits. One potential source of future work is to test the robustness of the subcircuits that we identified by perturbing the weights of the network itself. This way, we can find a subcircuit that is minimax-optimal. 

Moreover, our work operationalizes algorithmic changes as structure changes. We are the first to propose clustering techniques to quantify these structural changes. However, there may exist other methods of quantifying this change. We leave this open, and it could serve as a good basis for future work, especially in the mechanistic interpretability subfield. 

Lastly, our results could have strong implications for improving and explaining the logical reasoning abilities of transformers in general. We demonstrated that even on this simple task, language models are unable to identify and learn generalizable algorithms. Perhaps this is a barrier for more complex logical reasoning as well. 

\section{Conclusion}
In this paper, we demonstrate that for two-operand addition, surprisingly, language models are algorithmically unstable. Specifically, the underlying algorithm that is used to solve the problem changes as the task is perturbed. Our results are the first to look at algorithmic stability with respect to task perturbations. We believe that this interpretability technique will allow researchers to better evaluate language model design choices. Moreover, it may also serve as a diagnostic tool to determine whether or not generalizing behavior is occurring. We find that algorithmic instability correlates with a lack of generalizability. We hope that these insights will bridge many of the toy examples in mechanistic interpretability to practitioners.

\bibliography{references}

\newpage
\appendix

\section{Circuits}\label{sec:circuit-additional}
\raggedbottom
\begin{figure}[H]
    \centering
    \includegraphics[width=0.8\linewidth]{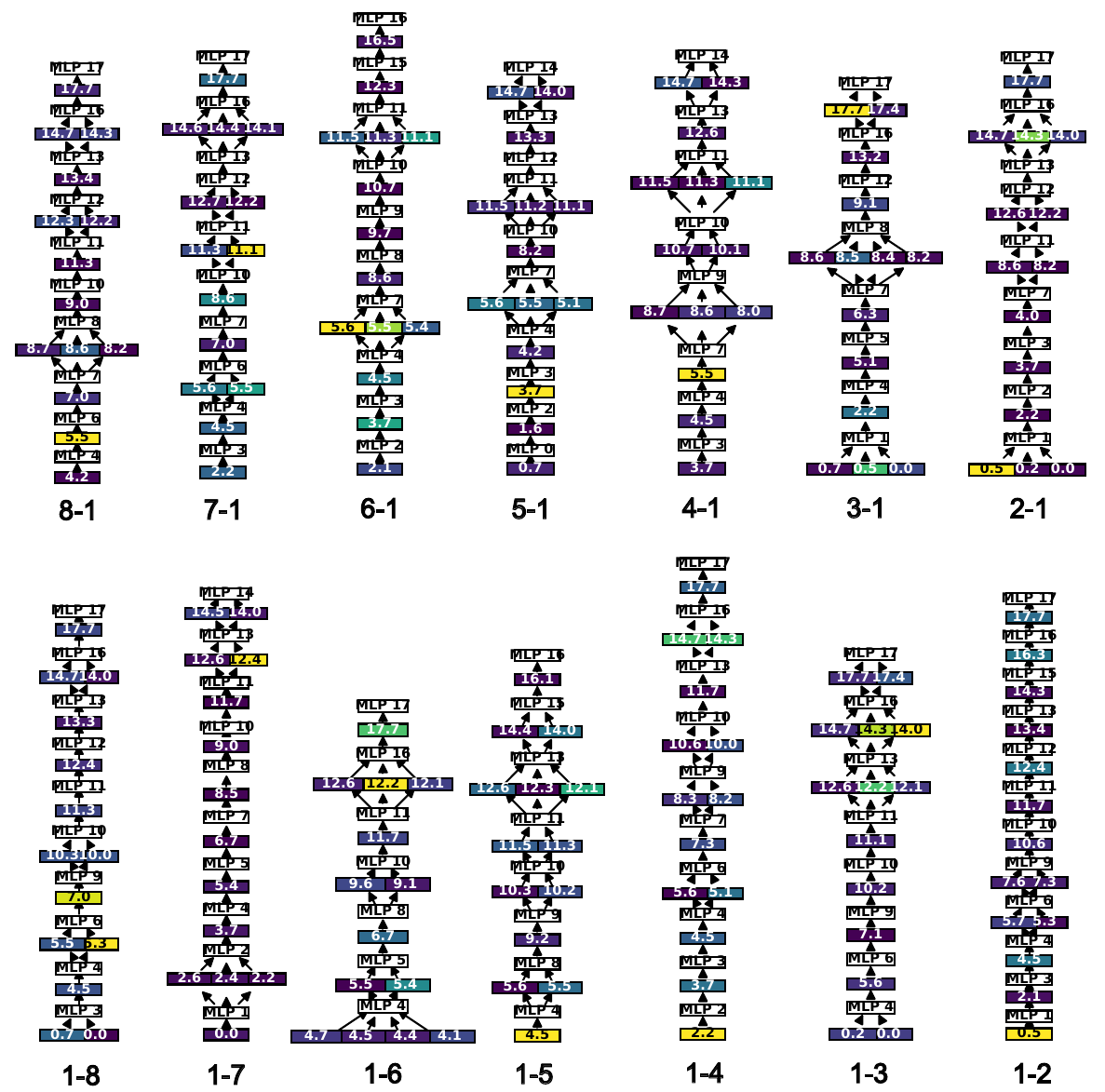}
    \caption{Circuits found for subtasks on the boundary. The attention heads shown are the top 10\% of the most influential heads with respect to our patching metric. Since we do not patch any of the MLP layers, some of them are simply omitted from the graphs for brevity.}
    \label{fig:boundary-circuits}
\end{figure}

\begin{figure}[H]
    \centering
    \includegraphics[width=0.8\linewidth]{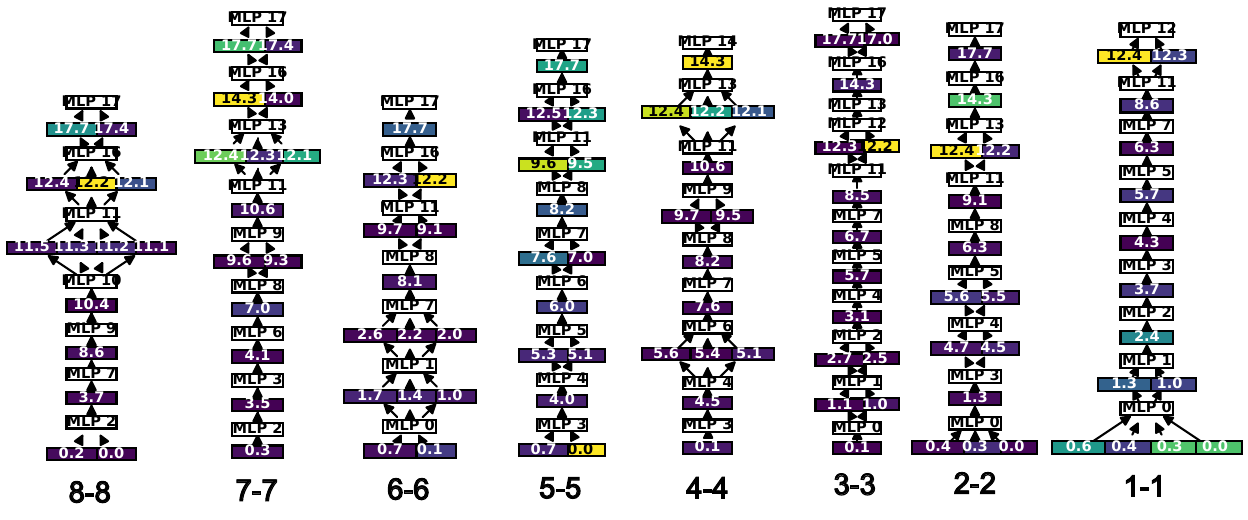}
    \caption{Circuits for the subtasks that are considered to be symmetric. The attention heads are the top 10\% of the most influential heads with respect to our patching metric. Between any two attention heads from different layers, if there are not other influential attention heads between them we omit showing all of the MLPs between them for brevity sake. However, we do not patch or remove the MLPs.}
    \label{fig:symmetric-circuits}
\end{figure}

\end{document}